\documentclass[letterpaper, 10 pt, conference]{ieeeconf}  

\IEEEoverridecommandlockouts                              

\usepackage{graphics} 
\usepackage{epsfig} 
\usepackage{mathptmx} 
\usepackage{times} 
\usepackage{amsmath} 
\usepackage{amssymb}  
\usepackage{pifont}
\usepackage{subcaption}
\usepackage{adjustbox}
\usepackage{array}
\usepackage{multirow, booktabs}
\usepackage{makecell}
\usepackage[mathscr]{euscript}
\usepackage{float}
\usepackage{gensymb}
\usepackage{mathtools}
\usepackage[bookmarks=true, hidelinks]{hyperref}    
\hypersetup{
     colorlinks = true,
     urlcolor = black
}

\usepackage{algpseudocode}
\usepackage{cite}
\usepackage{booktabs}

\title{\LARGE \bf
Mono-STAR: Mono-camera Scene-level Tracking and Reconstruction
}

\author{Haonan Chang$^{1}$, Dhruv Metha Ramesh$^{1}$, Shijie Geng$^{1}$, Yuqiu Gan, Abdeslam Boularias$^{1}$
\thanks{$^{1}$ Authors are with the Department of Computer Science,
        Rutgers University, 08854 New Brunswick, USA. This work is supported by NSF awards 1734492, 1846043, and 2132972.}%
}

\begin{document}

\maketitle
\thispagestyle{empty}
\pagestyle{empty}

\begin{abstract}
We present \emph{Mono-STAR}, the first real-time 3D reconstruction system that simultaneously supports semantic fusion, fast motion tracking, non-rigid object deformation, and topological change under a unified framework. 
The proposed system solves a new optimization problem incorporating optical-flow-based 2D constraints to deal with fast motion and a novel semantic-aware deformation graph (SAD-graph) for handling topology change. We test the proposed system under various challenging scenes and demonstrate that it significantly outperforms existing state-of-the-art methods.
Supplementary material, including videos, can be found at \href{https://github.com/changhaonan/Mono-STAR-demo}{\texttt{\textcolor{blue}{https://github.com/changhaonan/Mono-STAR-demo}}}.
\end{abstract}

\vspace{-0.5cm}
\section{INTRODUCTION}
Real-time perception is a crucial component of modern robotic manipulation systems. Recently, 
\textit{You Demonstrate Only Once}~\cite{wen2022you} has demonstrated that given the geometry model and 6D-pose trajectory of a manipulated object during an expert demonstration, a robot can quickly learn complex and contact-rich manipulation skills. Such progress shows the importance of geometric 3D reconstruction and tracking systems for robotic manipulation.

However, a perception system that can perform both tracking and reconstruction simultaneously is notoriously difficult to build because reconstruction and tracking inherently depend on each other. For example, tracking algorithms usually require geometry models, while dynamic scene reconstruction relies on accurate tracking for producing those geometry models. Scene-level Tracking and Reconstruction (\emph{STAR})~\cite{Haonan2022} refers to a category of perception systems that generate both the geometry and the pose of every visible object in a scene.

This problem is related to the multiple-instance dynamic SLAM problem, where all movable objects in the scene are assumed to be rigid so that the problem can be decomposed into multiple dense-SLAM sub-problems. This approach was proposed in {\it Co-Fusion}~\cite{Runz2017} and {\it MaskFusion}~\cite{Runz2019}, where a semantic neural network was employed first to decompose the scene into multiple objects and then deal with each object individually. This approach requires every object in the scene to be rigid or quasi-rigid. The same problem was investigated in {\it MidFusion}~\cite{xu2019mid}, where an octree was used to improve reconstruction and tracking. However, these systems  are limited to scenes of rigid objects with slow motions.

Instead of dealing with each object individually based on their semantic labels, one can also reconstruct all the objects in the scene as one large non-rigid object and segment them ulteriorly. This approach was however very challenging to apply until the introduction of  the first real-time non-rigid reconstruction {\it DynamicFusion} \cite{Newcombe}, where the non-rigid reconstruction problem was decomposed into two sub-problems, (1) building the geometry at the initial frame, and (2) computing the deformation using an embedded deformation graph, namely {\it ED-graph}. This paradigm was also followed in {\it OcclusionFusion}~\cite{lin2022occlusionfusion}.
Inspired by these previous efforts, a solution to the general STAR problem was recently proposed in {\it STAR-no-prior}~\cite{Haonan2022}. In contrast to  SLAM-based methods, {\it STAR-no-prior} reverses the order of segmentation and reconstruction. The entire scene is first reconstructed and then segmented into different objects based on topology. By doing so, {\it STAR-no-prior} outperforms previous state-of-the-art methods such as~\cite{Runz2019} and  MidFusion~\cite{xu2019mid}.

However, a major limitation of {\it STAR-no-prior} is its reliance on a  system of multiple cameras surrounding the scene, making it impractical for a mobile robot. To address this shortcoming, we propose \textbf{Mono-STAR}, a mono-camera STAR solution. Switching from a multi-camera system to a mono-camera setting requires solving several non-trivial problems. Notably,  {\it STAR-no-prior} relies on the multi-camera system to overcome 
the plane-based-ICP constraint that it inherits from {\it DynamicFusion}, which supports tracking of only slow motion along the camera view.
 The use of multiple cameras can guarantee that any motion has at least one non-zero projection to a camera view. However, the mono-camera setting does not have such a guarantee and therefore requires a new solution. {\it Occlusion Fusion}~\cite{lin2022occlusionfusion} adds a 2D constraint using optical-flow ({\it RAFT}~\cite{raft} or {\it GMA}~\cite{gma}) to deal with fast motions. Inspired by this, we propose a new 2D loss to track motions that are perpendicular to the camera view, which not only stabilizes tracking performance under a single view but also improves our system's ability to handle fast motion.

Furthermore, {\it STAR-no-prior} does not take advantage of semantic labels. We, therefore, combine the semantic information with the embedded deformation graph mechanism and propose a Semantic-aware Adaptive Deformation graph, {\it SAD-graph}, which is an extension of ED-graph. With just little extra computation, SAD-graph can easily handle topology changes across distinct semantic classes and assign different levels of rigidness for each type of object. To the best of our knowledge, Mono-STAR is the first single-view real-time 3D reconstruction system that can simultaneously handle semantic fusion, fast motion tracking, non-rigid object deformation, and topological change under one unified framework.

\begin{table}[ht!]
\centering
\begin{adjustbox}{width=\linewidth}
\begin{tabular}{cccccc}
\toprule
\multirow{2}{*}{Method} & Semantic & Fast & Non-rigid & Topology & Single\\
& & motion & objects & change& view\\ 
\cmidrule{1-6}
SLAM++  & \color{green}\ding{52}  & \color{red}\ding{56} & \color{red}\ding{56}   &\color{red}\ding{56}   &\color{green}\ding{52}\\ 
\specialrule{0em}{0.5pt}{0.5pt}
DynamicFusion \cite{Newcombe}  & \color{red}\ding{56} &  \color{red}\ding{56}  & \color{green}\ding{52}  &  \color{red}\ding{56} &  \color{green}\ding{52}   \\ 
\specialrule{0em}{0.5pt}{0.5pt}
Volume Deform  & \color{red}\ding{56} &  \color{red}\ding{56}  & \color{green}\ding{52}  &  \color{red}\ding{56} &  \color{green}\ding{52}   \\
\specialrule{0em}{0.5pt}{0.5pt}
SurfelWarp \cite{Gao2019}& \color{red}\ding{56}&  \color{red}\ding{56}& \color{green}\ding{52}&  \color{red}\ding{56}& \color{green}\ding{52}\\  
\specialrule{0em}{0.5pt}{0.5pt}
TCAFusion \cite{Li2020a} &  \color{red}\ding{56}&  \color{red}\ding{56}& \color{green}\ding{52}&  \color{green}\ding{52}& \color{red}\ding{56}\\ 
\specialrule{0em}{0.5pt}{0.5pt}
Co-fusion \cite{Runz2017} & \color{green}\ding{52} & \color{red}\ding{56}& \color{red}\ding{56}& \color{green}\ding{52}& \color{green}\ding{52} \\ 
\specialrule{0em}{0.5pt}{0.5pt}
\specialrule{0em}{0.5pt}{0.5pt}
Fusion4D \cite{Dou2016}& \color{red}\ding{56}&  \color{green}\ding{52}& \color{green}\ding{52}&  \color{green}\ding{52}& \color{red}\ding{56}\\ 
\specialrule{0em}{0.5pt}{0.5pt}
Motion2Fusion \cite{Dou2017} & \color{red}\ding{56}&  \color{green}\ding{52}& \color{green}\ding{52}&  \color{green}\ding{52}& \color{red}\ding{56}\\ 
\specialrule{0em}{0.5pt}{0.5pt}
Functon4D \cite{Yu}& \color{red}\ding{56}&  \color{green}\ding{52}& \color{green}\ding{52}&  \color{green}\ding{52}& \color{red}\ding{56}\\
MaskFusion \cite{Runz2019}& \color{green}\ding{52} & \color{red}\ding{56}& \color{red}\ding{56}& \color{green}\ding{52}& \color{green}\ding{52}\\ 
\specialrule{0em}{0.5pt}{0.5pt}
RigidFusion \cite{Wong2021}  & \color{green}\ding{52} & \color{red}\ding{56}& \color{red}\ding{56}& \color{green}\ding{52}& \color{green}\ding{52}\\
\specialrule{0em}{0.5pt}{0.5pt}
MidFusion \cite{xu2019mid}  & \color{green}\ding{52} & \color{red}\ding{56}& \color{red}\ding{56}& \color{green}\ding{52}& \color{green}\ding{52}\\ 
\specialrule{0em}{0.5pt}{0.5pt}
OcclusionFusion~\cite{lin2022occlusionfusion}  & \color{red}\ding{56} & \color{green}\ding{52} & \color{green}\ding{52}  &\color{red}\ding{56} & \color{green}\ding{52}\\ 
\specialrule{0em}{0.5pt}{0.5pt}
STAR-no-prior~\cite{Haonan2022}  & \color{red}\ding{56} & \color{red}\ding{56} & \color{green}\ding{52}  &\color{green}\ding{52} & \color{red}\ding{56}\\ 
\specialrule{0em}{0.5pt}{0.5pt}
\textbf{Mono-STAR}  & \color{green}\ding{52} & \color{green}\ding{52} & \color{green}\ding{52}  &\color{green}\ding{52} & \color{green}\ding{52}\\ 
\bottomrule
\end{tabular}
\end{adjustbox}
\caption{\small Taxonomy of the-state-of-art scene-level fusion systems.}
\label{taxonomy}
\vspace{-0.5cm}
\end{table}

\section{RELATED WORKS}

{\bf Simultaneous Tracking and Reconstruction.}
Simultaneous 6D tracking and 3D reconstruction was typically regarded in previous works as a multiple-instance dynamic SLAM problem. Many works such as Co-fusion~\cite{Runz2017}, MaskFusion~\cite{Runz2019}, and RigidFusion~\cite{Wong2021} proposed to divide the scene into multiple rigid objects and track each object individually. More recently, {\it STAR-no-prior}~\cite{Haonan2022} formalized the STAR problem as a scene-level non-rigid reconstruction problem. Our mono-camera system eliminates the multi-camera requirement of {\it STAR-no-prior} by adding a new optical-flow-based 2D constraint and a novel semantic-aware adaptive deformation graph.

{\bf Dynamic Scene Reconstruction.}
Dynamic scene reconstruction~\cite{Collet,dou20153d} is the problem of reconstructing the geometry and recording the deformation of a scene with moving objects. {\it DynamicFusion}~\cite{Newcombe} was the first real-time GPU-based solution for solving this problem. It adopts a TSDF-based geometry as the canonical model and an embed-deformation graph (ED-graph) to describe the deformation of the whole scene. A drawback of this method is that the combination of TSDF and ED-graph cannot handle topology changes. Many recent techniques such as {\it Fusion4D}~\cite{Dou2016}, {\it Motion2Fusion}~\cite{Dou2017}, \cite{Li2020a,Zampogiannis2018} have attempted to address this problem. However, these methods require significantly more computation or rely on expensive sensors. {\it SurfelWarp}~\cite{Gao2019} demonstrated that a Surfel-based representation can be used to tackle topology changes. Therefore, our proposed system also adopts a Surfel-based representation.
\section{problem formulation and background}
\subsection{Problem formulation}  \label{problem}
Given a sequence of RGB-D images of a given dynamic scene taken from a single fixed camera,
we consider the problem of simultaneous tracking and reconstruction of all the objects visible in the scene. The number of objects is unknown. The objects can be non-rigid. 
Measurement, $M_t$ can defined as set of measurement surfels $m_{i}$ at time-step $t$, generated from the RGB-D input. $m_i = (v_{i}, n_{i}, c_{i})$, where $v, n, c$ are 3D coordinates, normal and color respectively.

The proposed system returns at each time-step $t$ a {\it Surfel-based} geometry $S_t$ (the reconstructed scene) for the entire scene and its corresponding deformation graph $G_t$. Surfel-based geometry $S_t$ is a set of surfels $s_i$. $s_i = (v_i, n_i, c_i, r_i, l_{s_i})$, where $v_i, n_i, c_i, r_i, l_{s_i}$  are respectively the 3D coordinates, normal, color, radius and semantic label of surfel  $s_i \in S_t$.  We assume that there is a maximum of $H$  pre-defined different semantic categories $ \{1, 2, \ldots, H\}$. If a surfel does not belong to any pre-defined category, it will be labeled as $H+1$ (i.e., unrecognized). Deformation graph $G_t$ is defined by a set of nodes $\{g_i\}$. Each node $g_i$ has a semantic label $l_{g_i}$, and is connected to its nearest-neighbor nodes, denoted as $N^G(g_i)$, in the 3D space. Deformation graph $G_t$ is associated with a {\it warp field} $W_t$, defined as $W = \{ [p_i \in \mathbb{R}^3, \delta_i \in \mathbb{R}^+, T_i \in SE(3)]\}$, wherein $i$ is the index of a node in $G_t$, $p_i$ is the 3D point that corresponds to node $g_i$, $\delta_i$ is the node's radius of influence, and $T_i$ is the 6D transformation defined on node $g_i$.  $T_i$ is represented by a {\it dual quaternion} $q_i$ for smooth interpolation~\cite{Kavan2007}. Warp field $W$ describes the deformation between two consecutive time steps. For each surfel $s = (v, n, c, r, l) \in S$, we compute its 6D transformation $\bar W(s)$ based on warp field $W$,
\begin{equation}
\vspace{-0.1cm}
    \bar W(s) = normalize(\sum_{k\in N^G(s)} w(v,p_k)q_k),
    \label{warp_field}
    \vspace{-0.1cm}
\end{equation}
wherein $N^G(s)$ denotes the neighbors nodes of surfel $s$, $w(s)$ is an interpolation parameter, defined as $w(s) = \exp\big(\left\Vert  v - p_k \right\Vert^2_2 / (2\delta_k^2)\big)$, and $v$ is the 3D position of surfel $s$.
The local transformation $\bar W(s)$ is then used to describe the deformation of surfel $s$ as follows:
\begin{equation}
  \label{warp}
   v_{warp} = \bar W(s)v \; \quad  n_{warp} = \textrm{rotation}\big(\bar W(s)\big) n.
\end{equation}
Here, $v,n$ are the vertex and normal of $s$ before warping, and $v_{warp},n_{warp}$ are the vertex and normal after the deformation.

\begin{table}[ht]
\begin{tabular}{@{}ll@{}ll@{}}
\toprule
Symbol           & Meaning                   & Definition                                        \\ \midrule
$M_t$            & Measurement at time t. &~\ref{problem}, ~\ref{measurement}               \\
$S_{t-1}$        & Surfel geometry from t-1. &~\ref{problem}                 \\
$R^a_{t-1}$      & 2D maps rendered from $S_{t-1}$. &~\ref{rendering} \\
$S_{t-1}^{warp}$ & Warped geometry after non-rigid alignment. &~\ref{update_geometry}   \\
$R^g_{t-1}$      & 2D maps rendered from $S_{t-1}^{warp}$. &~\ref{rendering} \\ \bottomrule
\end{tabular}
\label{symbols}
\caption{Notation sheet.}
\vspace{-0.4cm}
\end{table}

\section{PROPOSED APPROACH}
\begin{figure*}[ht]
 \center
  \includegraphics[width=1.00\textwidth]{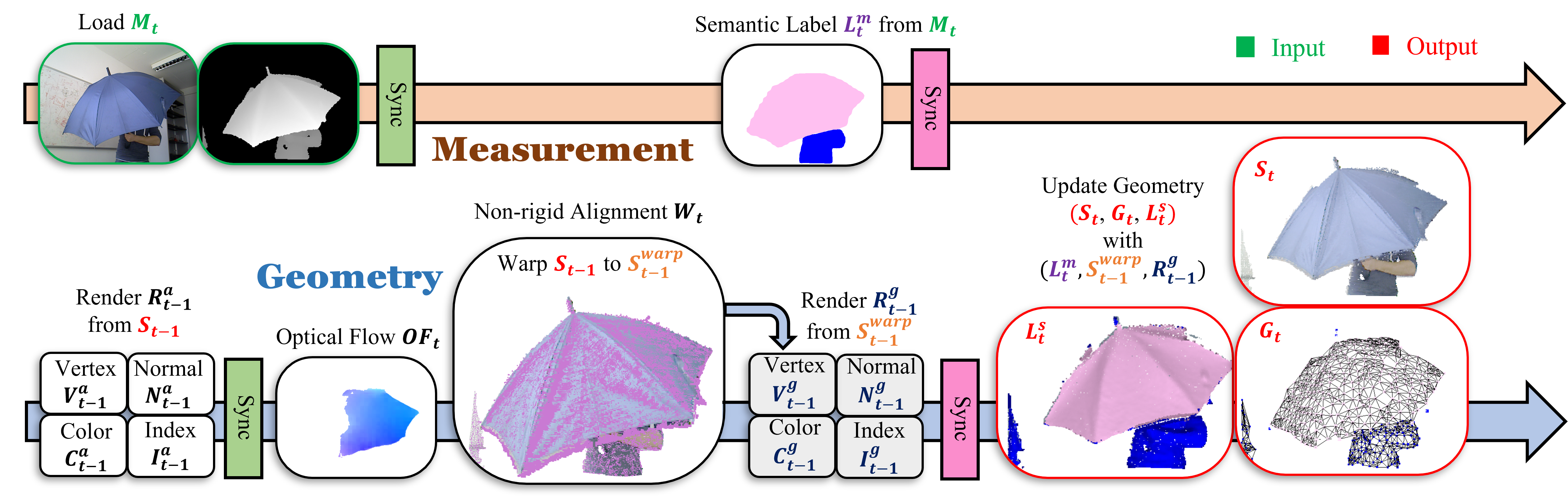}
  \caption{\small Overview of the proposed system. The system runs in two parallel threads, one for measurement and one for geometry. In each time-step $t$, the measurement thread loads a measurement $M_t$ from images or a camera buffer. Then, a segmentation network generates a set of semantic labels $L^m_{t}$. Once the measurement is loaded on the GPU memory, $M_t$ and previous alignment rendering $R^a_{t-1}$ are fed into an optical-flow network to generate the optical-flow $OF_t$ from previous geometry $S_{t-1}$ to measurement $M_t$.  Optical-flow $OF_t$, geometry rendering $R_t$ and measurement $M_t$ are used to compute warp-field $W_t$ with non-rigid alignment. After the alignment, previous geometry $S_{t-1}$ will be warped to $S_{t-1}^{warp}$. The fusion rendering map $R^g_{t-1}$ is then rendered from $S_{t-1}^{warp}$.  $R^g_{t-1}$, $S_{t-1}^{warp}$ and semantic labels $L^m_t$ are used to generate the updated geometry $S_t$, deformation graph $G_t$ and the surfel semantic label $L^s_t$. } 
  \label{Overview}
\vspace{-6pt}
\end{figure*}

An overview of the proposed method is shown in Fig.~\ref{Overview}. {\it Mono-STAR} uses two parallel threads, a measurement thread, and a geometry thread. The first thread is responsible for loading measurements $M_t$ and generating semantic labels $L^m_t$. The geometry thread uses this measurement $M_t$ and the alignment rendering $R^a_{t-1}$ to compute an optical-flow $OF_t$. Then, $M_t, R^a_{t-1}, OF_t$ are given to the optimization module that then computes the non-rigid deformation $W_t$. After the non-rigid alignment, previous geometry $S_{t-1}$ is warped to $S_{t-1}^{warp}$, and the geometry rendering  $R^{g}_{t-1}$ is generated. Finally, $R^{g}_{t-1}$, deformation $W_t$, warped geometry $S_{t-1}^{warp}$, and semantic labels $L^m_t$ are combined to generate the latest geometry $S_t$.

Noticeably, loading $M_t$ takes less time than updating geometry $S_{t-1}$ and rendering $R^a_{t-1}$. Semantic segmentation is also faster than the combined process of generating optical-flow $OF_t$ and the non-rigid alignment. Thus, the geometry thread fully hides the latency of the measurement thread.
\subsection{Measurement Thread}
\subsubsection{Measurement} \label{measurement}
We use one Intel RealSense-415 camera to collect RGB-D images. Depth images are denoised with a Gaussian filter. The maximum frame rate for this module is limited to $20$ fps to coordinate with other modules. We use a double-buffer strategy to hide latency. Specifically, we use two buffers $B_0$ and $B_1$ to store measurements. When $B_0$ is used by other threads, $B_1$ can read images simultaneously. The filtered images are used to construct three maps, $V_t^m$, $N_t^m$, $C_t^m$,  storing 3D coordinates $v_i^m$, normal $n_i^m$, and color $c_i^m$, respectively, for surfel $m_i$ of each pixel in the measurement.

\subsubsection{Segmentation}
The segmentation module receives the color map $C_t^m$ and returns a semantic label map $L_t^m$ of $H$ pre-defined semantic classes. Here, we use two different segmentation models, a transformer-based Segmenter Mask \cite{segmentor}, and a more traditional MaskRCNN \cite{maskrcnn}. The two models are pre-trained on two different datasets, ADE20K \cite{zhou2017scene, zhou2016semantic} and COCO-Stuff \cite{caesar2018cvpr} respectively. We do not further train these models on any other dataset. We select which one to use based on the types of objects in the scene.

\subsection{Geometry Thread}

\subsubsection{Geometry Rendering} \label{rendering}
The input of the geometry rendering pipeline is a geometry $S$, and the output is the rasterized rendering $R$ for geometry $S$ from the current camera view. The rendering process to generate $R$ follows the classical point cloud rasterizing process~\cite{pcd_raster}, where every surfel is projected to its nearest pixel position on the camera plane based on its 3D coordinates. Each rendering map $R$ is composed of four 2D maps.  $R_{t} = \{C_{t}, V_{t}, N_{t}, I_{t}\}$, where $V_{t}, N_{t}, C_{t}, I_{t}$ are respectively the vertex map, the normal map, the color map and the index map. These maps store the 3D coordinates $v_i$, the normal $n_i$, the color $c_i$ and the surfel index $i$ of the projected surfel $s_i$ at each pixel. 

At each time-step, the geometry rendering pipeline is called twice; once to generate $R^a_{t-1}$ from previous geometry $S_{t-1}$ for non-rigid alignment, and once to get $R^g_{t-1}$ with warped geometry $S^{warp}_{t-1}$ for  updating the geometry. Rendering $R_{t-1}^{g}$ used for updating the geometry operates on surfel-level granularity, whereas $R_{t-1}^{a}$ used for geometry alignment rendering operates on deformation node granularity.
Another difference between $R_{t-1}^{g}$ and $R_{t-1}^{a}$ is resolution, $R_{t-1}^{g}$ is up-sampled by $4 \times 4$ compared to $R_{t-1}^a$ to prevent different surfels from being projected onto the same pixel. $R_{t-1}^{g}$ requires a higher resolution for accurate geometry update and $R_{t-1}^{a}$ has a lower resolution for faster optimization.

\subsubsection{Optical Flow}
The optical flow module receives $C^a_{t-1}, V^a_{t-1}$ from geometry rendering $R^a_{t-1}$, and $C^m_{t-1}, V^m_{t-1}$ from measurement $M_t$, and generates an optical-flow map $OF_t$. $OF_t$ predicts the optical-flow from previous geometry $S_{t-1}$ to the latest measurement $M_{t}$. This prediction is later used for registration through non-rigid alignment. We generate $OF_{t}$ using a neural network based on the RAFT architecture~\cite{raft}, along with additional global motion features as performed in GMA~\cite{gma}. The global motion features provide stability for predicting motion features, even in occluded scenes. Both RAFT and GMA models were originally trained using only RGB images. The optical flow model used in~\cite{lin2022occlusionfusion} shows that using RGB-D images for training provides a far more stable flow, even with motion blurring. Thus, our model is also trained on RGB-D images from the datasets FlyingThings3D~\cite{MIFDB16}, Monkaa~\cite{MIFDB16} and Sintel~\cite{Butler:ECCV:2012, Wulff:ECCVws:2012}.

\subsubsection{Non-rigid alignment}
Non-rigid alignment is performed in order to compute non-rigid deformation $W_t$. This step solves a massive optimization problem to warp the previous geometry $S_{t-1}$ to a geometry $S_{t-1}^{warp}$ that fits current measurement $M_t$. We use a Gauss-Seidel solver implemented with CUDA to solve this problem, which is summarized as
\begin{align*}
     min_{W}E_{total}(W) = w_{picp} E_{picp}(W) + & w_{2D} E_{2D}(W) \\ + & w_{areg} E_{areg}(W),
\end{align*}
where $w_{picp}, w_{of}, w_{areg}$ are the weights of terms $ E_{picp}(W)$, $E_{2D}(W)$ and $E_{areg}(W)$, explained in the following. 

{\bf Registration.} Let $u=(x, y)$ be a pixel in measurement map $u$, and let $m_i = M(u)$ bet its associated surfel. Let $(m_i, s_{\Pi(i)})$ denote a pair of registered measurement and geometry surfel. $\Pi(i)$ is defined as   $\Pi(i) \coloneqq I^a_{t-1}(x - of^x_t, y - of^y_t)$, wherein $(x, y)=u, (of^x_t, of^y_t) = OF_t(u)$. $I_{t-1}^a \in R_{t-1}^a$ is the index map of the rendered geometry.

{\bf PICP Loss.} Point-to-point ICP loss is sensitive to disturbance and outliers, which limits its utility in real-world applications. Instead, we use a plane-based ICP (PICP) loss to align the differences along the depth direction as follows,
\begin{equation}
    E_{picp}(W) = \sum_{m_i \in M} n_i^m \cdot (\Bar{W}(s_{\Pi(i)})v^s_{\Pi(i)} - v_i^m),
\end{equation}
wherein $v^s_{\Pi(i)}$ is the 3D coordinates of surfel $s_{\Pi(i)}$, $v_i^m, n_i^m$ are the 3D coordinates and normal of measurement surfel $m_i$. $\Bar{W}$ is defined in Eq.~\ref{warp}. 

{\bf 2D Loss.} One limitation of the PICP loss is that it cannot correctly capture motions within the same plane, such as the moving calendar shown in Fig.~\ref{ablation_of}. We thus add to the objective function a 2D loss $E_{2D}$ defined as follows,
\begin{align}
P & = 
\begin{pmatrix}
1 & 0  & 0 \\
0 &1 & 0
\end{pmatrix} \\
E_{2D}(W) & = \sum_{m_i \in M} \lVert P (\Bar{W}(s_{\Pi(i)})v^s_{\Pi(i)} - v_i^m) \rVert _2.
\end{align}
Here, $P$ is a projection matrix, projecting the 3D difference to the camera X-Y plane. This term constrains $m_i$ and $s_{\Pi(i)}$ to be as close as possible on the camera X-Y plane. It is worth noting that our proposed 2D loss is different from the one proposed in OcclusionFusion~\cite{lin2022occlusionfusion}, where pixel differences are used to calculate the 2D loss. The influence of pixel differences scales with the distance to the camera, which makes the optimization parameters harder to tune. 

\begin{figure}[ht]
 \center
  \includegraphics[width=0.48\textwidth]{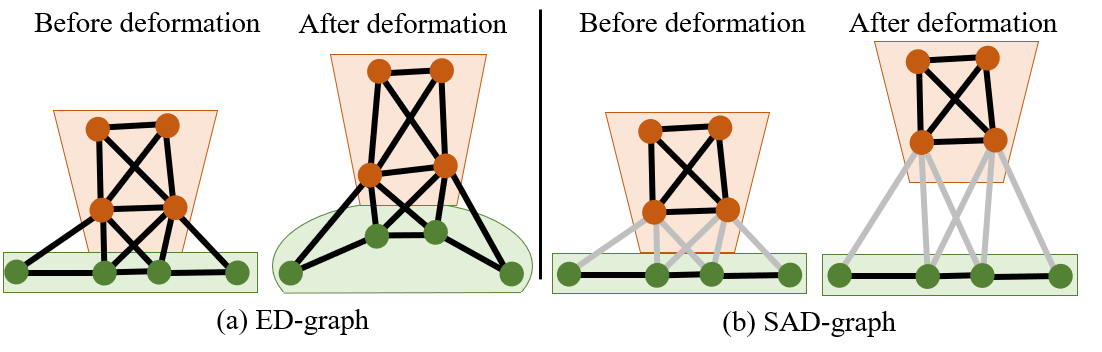}
  \caption{\small Illustration of the proposed Semantic-aware Adaptive Deformation Graph (SAD-graph). The scene describes a cup (brown) being lifted up from the table (green). Black edges indicate strong continuity constraints, while grey edges indicate weak constraints. }
  \label{sad_graph}
\end{figure}

{\bf Semantic-aware Adaptive Deformation Graph.} \label{sec_sad_graph}
The traditional {\it Embedded Deformation} graph (ED-graph) has been widely used in non-rigid tracking and non-rigid reconstruction. It can describe complicated warping fields with a simple data structure and an interpolation strategy. Moreover, the {\it as-rigid-as-possible} (ASAP) regulation term defined on deformation nodes provides a continuity guarantee for neighboring nodes. However, the ED-graph cannot handle the topology changes of different nodes. For example, if we use an ED graph to describe a cup being lifted up from a table, as shown in Fig.~\ref{sad_graph} (a), the motion of the cup's nodes also influences and propagates to the table's nodes. Many previous works have attempted to overcome this limitation of the ED-graph by proposing a dual deformation graph \cite{Zampogiannis2018} or a level-set-based TSDF fusion mechanism \cite{Li2020a}. However, these approaches are too complicated or introduce too much overhead computation. 
We propose the {\it Semantic-aware Adaptive Deformation Graph} (SAD-graph) to address this issue of topological changes among objects with different semantic classes. More importantly, the proposed algorithm is intuitive and requires little extra computation compared with ED-graph. Another advantage is that existing ED-graph based approaches can be easily upgraded to support SAD-graph.

The core idea of SAD-graph is that instead of imposing a uniform regulation continuity constraint on all deformation nodes, constraints of varying adaptive strengths are imposed on different edges. A variable weight $\omega_{i, j}$ is associated with the constraint (defined in Eq.~\ref{adaptive_reg_eq}) between neighboring nodes $g_i$ and $g_j$, and the strength of the constraint is systematically adjusted. Constraint weight $\omega_{g_i, g_j}$ is a function of  $l_{g_i}, l_{g_j}$, the semantic labels of nodes $g_i$ and $g_j$. It is defined as follows,
\begin{equation}
\label{connection_weight}
    \omega_{g_i, g_j}(l_{g_i}, l_{g_j}) = \begin{cases}
    0.1 ,& \text{if } l_{g_i} \neq l_{g_j}\\
    \delta^k,& \text{if } l_{g_i} = l_{g_j} = k, k \in [1, \ldots, H+1]
\end{cases} 
\end{equation}
where $\delta^k$ is a constant describing the average rigidness of objects belonging to semantic category $k$, e.g., $\delta^{table}=1.0$, $\delta^{human}=0.3$, etc. For example,  in Fig.~\ref{sad_graph} (b), since the internal rigidity constraint within the cup or the table is much larger than the constraint between them, their geometries can be accurately reconstructed during the topology separation. It is important to note that these constraints are not hard because the semantic labels obtained from a neural network detector are error-prone. 

{\bf Adaptive Regulation Loss.} We introduce a deformation graph, SAD-graph, and a new regulation term, {\it adaptive regulation} $E_{areg}(W)$. A semantic-related connection weight $\omega$ is used to adjust the regulation strength among and within different semantic classes as follows,
\begin{equation}
    E_{areg}(W) = \sum_{g_j \in G} \sum_{g_i \in N^G(g_j)} \omega_{g_i, g_j} \left\Vert  T_j p_j - T_i p_i \right\Vert^2_2, 
    \label{adaptive_reg_eq}
\end{equation}
wherein $G$ is the deformation graph, $N^G(g_j)$ refers to the set of neighbors of node $g_j$ in the deformation graph, $T_j$ and $T_i$ are the transformations defined on nodes $g_i$ and $g_j$. $p_i$ and $p_j$ are the 3D coordinate of $g_i$ and $g_j$, and $\omega_{g_i, g_j}$ is the weight of the connection between nodes $g_i$ and $g_j$, defined in Eq.~\ref{connection_weight}.

\subsubsection{Geometry and Graph Update} \label{update_geometry}
Once the non-rigid deformation is computed, the geometry update process of Mono-STAR is similar to SurfelWarp~\cite{Gao2019}. Thus, we only briefly describe that process and we focus on the semantic update. This step returns the updated geometry $S_t$ (the reconstructed scene) and the updated graph $G_t$, both of which are needed for processing the scene in the next time-step.

{\bf Updating the Geometry.} The previous geometry $S_{t-1}$ is warped to $S_{t-1}^{warp}$ after the non-rigid alignment step. Although $S_{t-1}^{warp}$ is already close enough to measurement $M_t$, there still exists a discrepancy between them due to measurement noises, emerging surfaces, topology changes, or even tracking failures. The geometry is updated to address this gap between the warped geometry $S_{t-1}^{warp}$ and measurement $M_t$. There are four steps in total in this process. 

1. {\it Registration}: A projective registration is made between measurement $M_t$ and warped geometry $S_{t-1}^{warp}$ according to rendering map $R^{g}_{t-1}$. 

2. {\it  Fusion}: If a surfel $m_i \in M_t$ is mapped to  $s_j \in S_{t-1}^{warp}$ in the registration, $m_i$ is merged into $s_j$ to average measurement noises. The semantic label $l_{s_j}$ of $s_j$ is defined as a probability distribution $p_{s_j}$. When $m_i$ is fused into $s_j$, $p_{s_j}$ is also updated by $l_{m_i}$. The update formula for $p_{s_j}$ is:
\begin{align}
    p_{s_j}(k) = (p_{s_j}(k) + \delta_m)/\sum_{k'} {p_{s_j}(k')}, \text{if } k = l_{m_i} \\
    p_{s_j}(k) = p_{s_j}(k)/\sum_{k'} {p_{s_j}}(k'), \text{otherwise}.
\end{align}
Here, $\delta_m$ is the confidence of the measurement.

3. {\it Append}: If there are no surfels in $S_{t-1}^{warp}$ that can be registered to $m_i$, $m_i$ must belong to a newly observed surface or be noise. In the first case, $m_i$ will be appended to $S_{t-1}^{warp}$. The semantic label distribution $p^s_i$ of $m_i$ is initialized as:
\begin{equation}
    p^s_i(k) = \delta_m, \text{if } k = l_{m_i};  p^s_i(k) = 0, \text{otherwise}, 
\end{equation}

4. {\it Removal}: After each $m_i \in M_t$ is either fused or appended, some surfels $s_j \in S_{t-1}^{warp}$ are left with no correspondence. A geometry violation test is performed on the remaining surfels, and those that fail the test are removed~\cite{Gao2019}.

After the four steps given above, we get the updated geometry  $S_t$ for time-step $t$.

{\bf Updating the Graph.} 
The update of the SAD-graph is identical to the update of the traditional ED-graph. The update appends new nodes but does not remove existing ones.
Let $S^{append}$ be the set of the appended surfels during the geometry update. We first compute the distances between every surfel $s \in S^{append}$ and every node $g \in G_{t-1}$. Let $D(s, G) = min_{g\in G}  \textrm{ distance(s, g)}$. A surfel $s$ is said to be unsupported if $D(s, G) > \sigma$, for some threshold $\sigma$. We perform a spatially uniform sampling from all the unsupported surfels. Sampled surfels are appended to graph $G$ as new nodes. The semantic label of node $g_i$, $l_{g_i}$ is updated according to  the semantic labels of $N^S(g_i)$, neighbor surfels of $g_i$. $l_{g_i} = argmax_k\{\sum_{s_j\in N^s(g_i)}\delta(l_{s_j}, k)\}$. Here $\delta(l_{s_j}, k) = 1, \text{if } l_{s_j} = k;  \delta(l^s_j, k) = 0, \text{otherwise}$.
\section{EXPERIMENTS}
 We test our technique on a dataset we collected and a public dataset VolumeDeform~\cite{Innmann2016}. An ablation study and comparisons with SoTA methods such as STAR-no-prior~\cite{Haonan2022} and MaskFusion~\cite{Runz2019}  on challenging scenes are presented in this section. Since collecting ground-truth geometry and deformation for non-rigid objects is extremely challenging, experiments and comparisons in this area are limited to qualitative results~\cite{Innmann2016}. Supplementary results and resources can be found at {\small \href{https://github.com/changhaonan/Mono-STAR-demo}{\texttt{\textcolor{blue}{https://github.com/changhaonan/Mono-STAR-demo}}}}.

\subsection{Performance}
We tested our system on a desktop machine with a GeForce RTX 3090 and an AMD-Ryzen 9 5900X. On average, measurement loading takes $4$ ms and segmentation costs 10 ms (Segmenter Mask \cite{segmentor}). The optimization module uses $20$ ms. The geometry update uses $7$ ms. The major bottleneck is the optical-flow network, which takes $60$ ms. Since the latency for the measurement thread is fully hidden by the geometry thread, our entire system  runs in {\bf 11 Hz}. If the optical-flow runs on a separate graphic card, it would take only $26$ ms~\cite{lin2022occlusionfusion}, which would double the speed of our system.

\subsection{Qualitative Results}
\subsubsection{Soft objects}
\begin{figure}[ht]
 \center
  \includegraphics[width=0.48\textwidth]{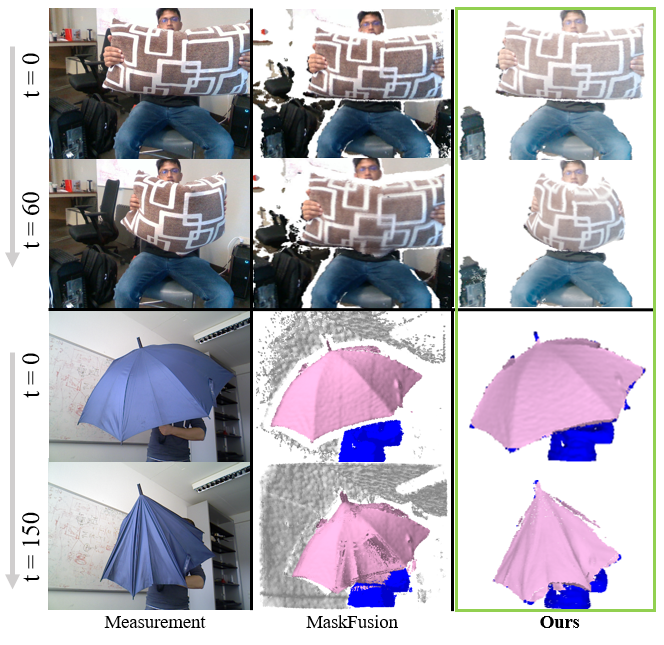}
  \caption{\small Comparison on deformable objects with MaskFusion on our recorded dataset (top) and VolumeDeform dataset (bottom). }
  \label{exp_deform}
\end{figure}

Fig.~\ref{exp_deform} illustrates the non-rigid deformation ability of MaskFusion and Mono-STAR. We can clearly see that MaskFusion fails to track the deformations of the pillow and umbrella, while Mono-STAR correctly captures both of them in the reconstructed model, which shows the advantage of our technique over MaskFusion in handling non-rigid deformation.

\subsubsection{Fast Motion}
\begin{figure}[ht]
 \center
  \includegraphics[width=0.48\textwidth]{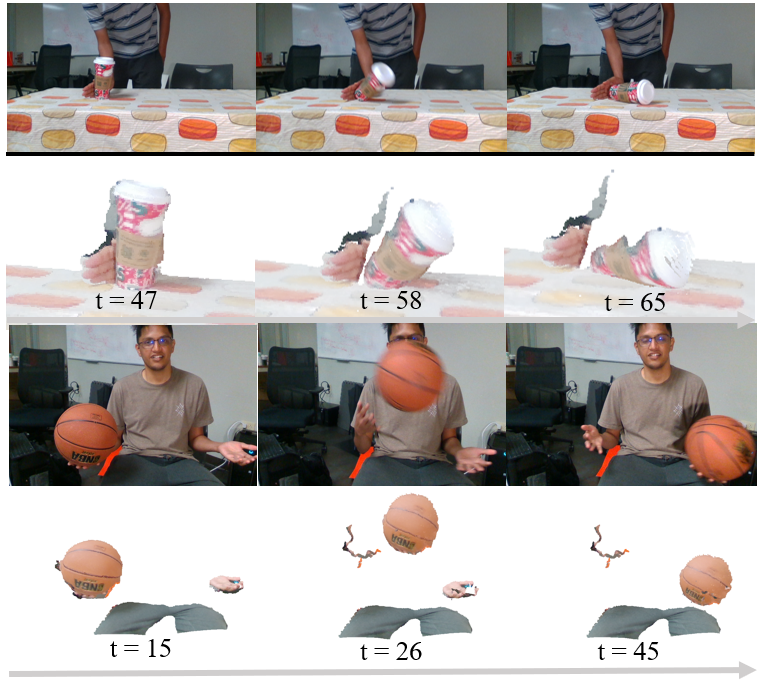}
  \caption{\small Experiment on fast motion. Pushing down a coffee cup (top). Passing a basketball between two hands (bottom). The second and fourth row are our 3D scene reconstruction results. }
  \label{coffee_cup}
\end{figure}

Fig.~\ref{coffee_cup} demonstrates Mono-STAR's ability to handle fast motions.
The top scene in Fig.~\ref{coffee_cup} shows an accident that was recorded during our data collection. While we were pushing a cup on the table, the cup hit a bump and fell down. The bottom scene is about passing a basketball between two hands. Objects in both scenes moved very fast. One is 18 frames, and the other is 30 frames. Significant motion blur can be observed in both middle images. However, Mono-STAR can still capture these fast motions and correctly reconstruct the objects at each frame.

\subsubsection{Resilience to Semantic Segmentation Noises}
\begin{figure}[ht]
 \center
  \includegraphics[width=0.48\textwidth]{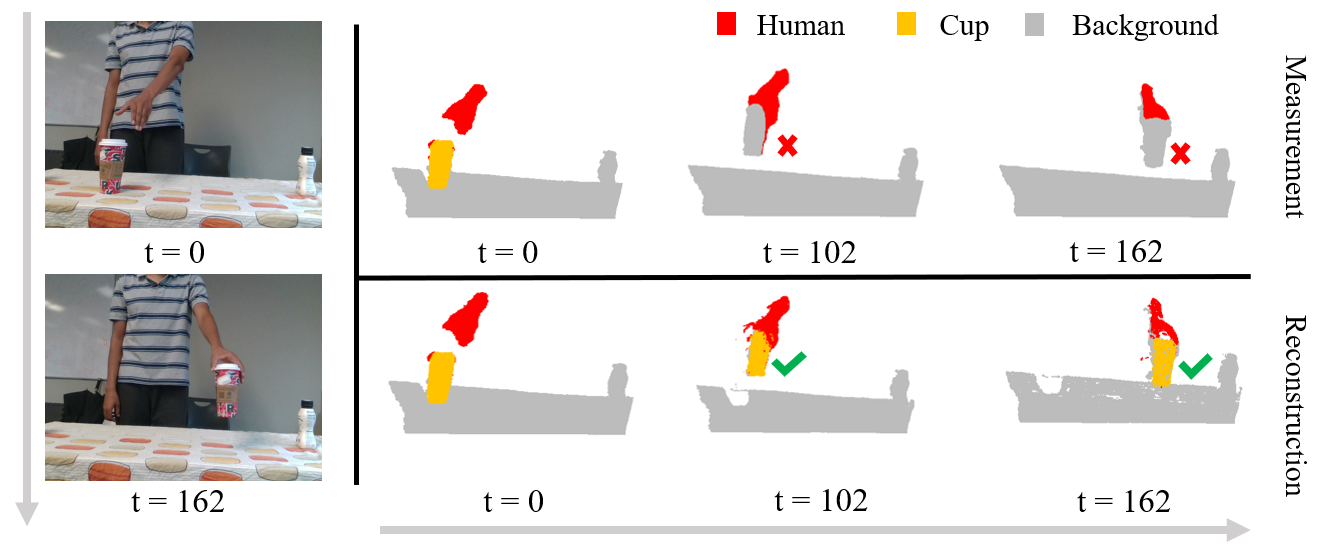}
  \caption{\small Resilience to semantic label noise. The top sequence is the segmentation map $L^m_t$ of the measurement. The bottom sequence is the segmentation map $L^s_t$ from our reconstruction technique.}
  \label{sementic_noise_resistant}
\end{figure}

Fig.~\ref{sementic_noise_resistant} shows how our proposed method can resist noise in semantic segmentation. The figures on the left are the RGB measurement from the beginning and the end frames. The right side compares the segmentation from the measurement and the segmentation from our reconstruction. Although the ground-truth measurement suffers from major segmentation errors, where the cup label is completely lost for $t>0$, Mono-STAR still maintains the correct semantic labels in its reconstruction result through semantic fusion.

\subsection{Ablation Study}
\subsubsection{2D Loss}
\begin{figure}[ht]
 \center
  \includegraphics[width=0.48\textwidth]{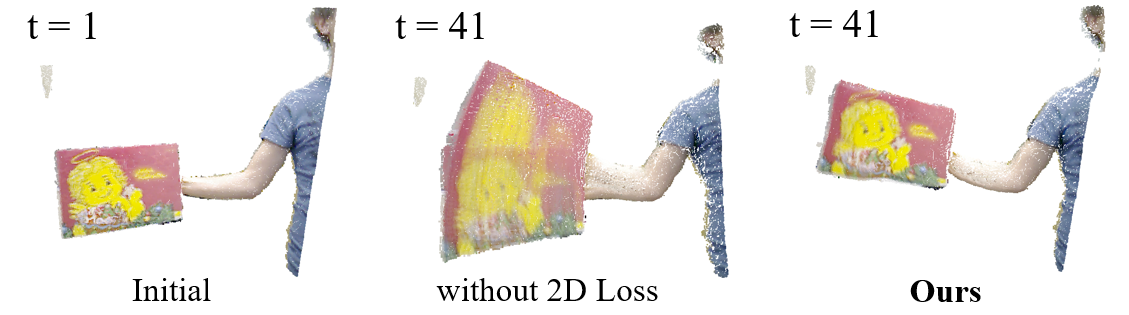}
  \caption{\small Ablation study on the 2D loss.}
  \label{ablation_of}
  \vspace{-0.5cm}
\end{figure}

We test Mono-STAR with and without the 2D loss on the ``adventcalender'' dataset from VolumeDeform. Fig.~\ref{ablation_of} shows that the proposed 2D loss $E_{2D}$ can efficiently track the motions within a plane. In contrast,  tracking without $E_{2D}$ fails in this type of motion, which clearly shows the effectiveness of the proposed 2D loss.

\subsubsection{SAD-graph}
\begin{figure}[ht]
 \center
  \includegraphics[width=0.48\textwidth]{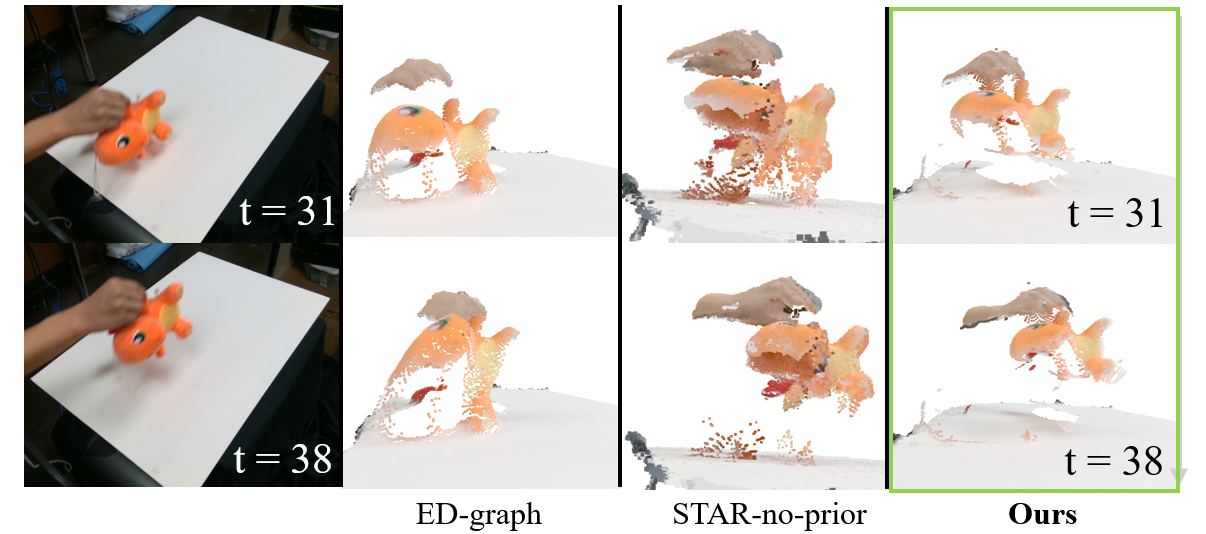}
  \caption{\small Comparing ED-graph (left), topology-aware ED-graph (middle, STAR-no-prior~\cite{Haonan2022}), and SAD-graph (right, ours). The scene shows a plushy toy being lifted up from the table.}
  \label{graph-comparison}
\end{figure}

In Fig.~\ref{graph-comparison}, we compare the ED-graph with the topology-aware ED-graph (STAR-no-prior) and the SAD-graph (Mono-STAR). We can see that the ED-graph fails to support the topology change that results from lifting the object from the table. Topology-aware ED-graph can separate the topology, but it also generates many outliers on the table. With the help of the proposed SAD-graph, Mono-STAR can conduct a smoother and cleaner separation.
  \vspace{-0.5cm}
\subsection{Discussion of Limitations}
Although Mono-STAR shows great potential in many different aspects, it still has two limitations. First, it relies on the optical flow to track fast motions. However, even the state-of-art optical flow detector GMA~\cite{gma} is not always accurate, especially when the motion is too fast and the tracked surfaces are heavily occluded. Our system can tolerate some noise from the GMA optical-flow module. However, if the optical flow provides inaccurate predictions for multiple consecutive frames, the tracking of the corresponding object may still fail. Another drawback of our system is the incompleteness of the reconstructed geometry. Our reconstructed geometries usually have holes and are not as smooth as TSDF-based geometry. The reason is that Surfel-based geometry, unlike TSDF-based geometry, is discrete by default. Therefore, it is difficult to maintain the smoothness of Surfel-based geometry in highly dynamic scenes. These two challenges can be addressed in future works.
\section{CONCLUSION}
We presented Mono-STAR, a single-view solution for the semantic-aware STAR problem. Mono-STAR uses a novel semantic-aware and adaptive deformation graph for simultaneous tracking and reconstruction, and can handle topology changes as well as semantic fusion. Experiments show that Mono-STAR achieves promising results in non-rigid object reconstruction, while resisting to semantic segmentation errors, and capturing fast motions on various challenging scenes. We believe that this system can inspire and boost more future research on imitation learning, dexterous manipulation, and many other relevant robotics problems.

\bibliographystyle{IEEEtran}
\bibliography{Star} 
\end{document}